\documentclass{amsart}
\newtheorem{thm}{Theorem}[section]
\newtheorem{cor}[thm]{Corollary}
\newtheorem{lem}[thm]{Lemma}

\theoremstyle{definition}

\theoremstyle{remark}

\newcommand{\norm}[1]{\left\Vert#1\right\Vert}
\newcommand{\Real}{\mathbb R}
\renewcommand{\b}[1]{\mathbf{#1}}
\newcommand{\bx}{\b{x}}
\newcommand{\by}{\b{y}}
\newcommand{\bu}{\b{u}}
\newcommand{\citep}[1]{(\cite{#1})}
\newcommand{\s}{\hskip1cm}

\begin{document}

\title[RL with LFA and LQ Control Converges]{Reinforcement Learning with Linear Function Approximation and LQ control Converges\footnotemark{*}}

\thanks{$^*$Last updated: 22 October 2006}%
\author[I. Szita and A. L\H{o}rincz]{Istv\'{a}n Szita \and Andr\'{a}s L\H{o}rincz}
\address{\newline Department of Information Systems \newline E\"{o}tv\"{o}s Lor\'{a}nd University of
Sciences \newline P\'azm\'any P\'eter s\'et\'any 1/C \newline 1117
Budapest, Hungary \newline Emails:
\newline Istv\'an Szita: szityu@stella.eotvos.elte.hu
\newline Andr\'as L{\H o}rincz: lorincz@inf.elte.hu
\newline WWW:
\newline \texttt{http://nipg.inf.elte.hu}
\newline \texttt{http://people.inf.elte.hu/lorincz}}

\begin{abstract}
Reinforcement learning is commonly used with function
approximation. However, very few positive results are known about
the convergence of function approximation based RL control
algorithms. In this paper we show that TD(0) and Sarsa(0) with
linear function approximation is convergent for a simple class of
problems, where the system is linear and the costs are quadratic
(the LQ control problem). Furthermore, we show that for systems
with Gaussian noise and non-completely observable states (the LQG
problem), the mentioned RL algorithms are still convergent, if
they are combined with Kalman filtering.
\end{abstract}

\maketitle

\section{Introduction}

Reinforcement learning is commonly used with function
approximation. However, the technique has little theoretical
performance guarantees: for example, it has been shown that even
linear function approximators (LFA) can diverge with such often
used algorithms as Q-learning or value iteration
\cite{baird95residual,sutton98reinforcement}. There are positive
results as well: it has been shown
\cite{tsitsiklis96analysis,precup01offpolicy,tadic01convergence}
that TD($\lambda$), Sarsa, importance-sampled Q-learning are
convergent with LFA, if the policy remains constant (policy
evaluation). However, to the best of our knowledge, the only
result about the control problem (when we try to find the optimal
policy) is the one of Gordon's \cite{gordon01reinforcement}, who
proved that TD(0) and Sarsa(0) can not diverge (although they may
oscillate around the optimum, as shown in
\cite{gordon96chattering})\footnote{These are results for policy
iteration (e.g. \cite{perkins02convergent}). However, by
construction, policy iteration could be very slow in practice.}.

In this paper, we show that  RL control with linear function
approximation can be convergent when it is applied to a linear
system, with quadratic cost functions (known as the LQ control
problem). Using the techniques of Gordon
\cite{gordon01reinforcement}, we were prove that under appropriate
conditions, TD(0) and Sarsa(0) converge to the optimal value
function. As a consequence, Kalman filtering with RL is convergent
for observable systems, too.

Although the LQ control task may seem simple, and there are
numerous other methods solving it, we think that this Technical
Report has some significance: (i) To our best knowledge, this is
the first paper showing the convergence of an RL control algorithm
using LFA. (ii) Many problems can be translated into LQ form
\cite{bradtke93reinforcement}.

\section{the LQ control problem}

Consider a linear  dynamical system with state $\b x_t \in
\Real^n$, control $\b u_t \in \Real^m$,
 in discrete time $t$:
\begin{eqnarray}
  \b x_{t+1} &=& F \b x_t + G \b u_t.
\end{eqnarray}
Executing control step $\b u_t$ in $\b x_t$ costs
\begin{equation}
c(\bx_t, \bu_t) := \bx_t^T Q \bx_t + \bu_t^T R \bu_t,
\end{equation}
and after the $N^{th}$ step the controller halts and receives a final cost of $\bx_N^T Q_N
\bx_N$. The task is to find a control sequence with minimum total cost.

First of all, we slightly  modify the problem: the run time of the
controller will not be a fixed number $N$. Instead, after each
time step, the process will be stopped with some fixed probability
$p$ (and then the controller incurs the final cost $c_f(\bx_f) :=
\bx_f^T Q^f \bx_f$). This modification is commonly used in the RL
literature; it makes the problem more amenable to mathematical
treatments.

\subsection{The cost-to-go function}

Let $V^*_t(\bx)$ be the optimal cost-to-go function at time step $t$, i.e.
\begin{equation} \label{e:V_original}
  V^*_t(\bx) := \inf_{\bu_t, \bu_{t+1}, \ldots} E\bigl[ c(\bx_t,\bu_t) + c(\bx_{t+1},\bu_{t+1}) +
  \ldots + c_f(\bx_f)  \big| \bx_t = \bx \bigr].
\end{equation}

Considering that the controller is stopped with probability $p$, Eq.~\ref{e:V_original} assumes
the following form
\begin{equation}
  V^*_t(\bx) = p \cdot c_f(\bx) + (1-p) \inf_{\bu} \Bigl( c(\bx,\bu) +
    V^*_{t+1} (F\bx + G\bu) \Bigr)
\end{equation}
for any state $\bx$. It is an easy matter to show that the optimal cost-to-go function is
time-independent and it is a quadratic function of $\bx$. That is, the optimal cost-to-go
action-value function assumes the form
\begin{equation}
  V^*(\bx) = \bx^T \Pi^* \bx.
\end{equation}

Our task is to estimate the optimal value functions (i.e., parameter matrix $\Pi^*$) on-line.
This can be done by the method of temporal differences.

We start with an arbitrary initial cost-to-go function $V_0(\bx) = \bx^T \Pi_0 \bx$. After this,
\begin{enumerate}
    \item control actions are selected according to the current value
function estimate
    \item the value function is updated according to the experience, and
    \item these two steps are iterated.
\end{enumerate}

The $t^{th}$ estimate of $V^*$ is $V_t(\bx) = \bx^T \Pi_t \bx$.
The greedy control action according to this is given by
\begin{eqnarray}
  \bu_t &=& \arg\min_{\bu} \Bigl( c(\bx_t,\bu)+ V_t(F \bx_t + G \bu) \Bigr) \label{u_t}\\
      &=& \arg\min_{\bu} \Bigl( \bu^T R \bu + (F \bx_t + G \bu)^T \Pi_t (F \bx_t + G \bu) \Bigr) \nonumber\\
      &=& -(R+G^T \Pi_t G)^{-1} (G^T \Pi_t F) \bx_t.  \nonumber
\end{eqnarray}

The 1-step TD error is
\begin{equation}
  \delta_t =
  \begin{cases}
    c_f(\bx_t) - V_t(\bx_{t})  & \text{if $t=t_{STOP}$}, \\
    \bigl( c(\bx_{t},\bu_{t}) + V_t(\bx_{t+1}) \bigr) - V_t(\bx_{t}), & \text{otherwise}.
  \end{cases}
\end{equation}

and the update rule for the parameter matrix $\Pi_t$ is
\begin{eqnarray}
  \Pi_{t+1} &=& \Pi_t + \alpha_t \cdot \delta_t \cdot \nabla_{\Pi_t} V_t(\bx_t) \label{Pi_update}\\
            &=& \Pi_t + \alpha_t \cdot \delta_t \cdot \bx_t \bx_t^T, \nonumber
\end{eqnarray}
where $\alpha_t$ is the learning rate.

The algorithm is summarized in Fig. \ref{LQ_TD}.

\begin{figure}
\begin{tabular}{l} 
\hline \hline
\\
Initialize $\bx_0$, $\bu_0$, $\Pi_0$ \\
repeat \\
\s $\bx_{t+1} = F \bx_t + G \bu_t$ \\
\s $\nu_{t+1}:=$ random noise \\
\s $\bu_{t+1} = - (R + G^T \Pi_{t+1} G)^{-1} (G^T \Pi_{t+1} F) \bx_{t+1}+\nu_{t+1}$ \\
\s with probability $p$, \\
\s \s $\delta_t =  \bx_t^T Q^f \bx_t - \bx_{t}^T \Pi_t \bx_{t} $\\
\s \s STOP \\
\s else \\
\s \s $\delta_t =  \bu_t^T R \bu_t + \bx_{t+1}^T \Pi_t \bx_{t+1} - \bx_{t}^T \Pi_t \bx_{t} $ \\
\s $\Pi_{t+1} =  \Pi_t + \alpha_t \delta_t \bx_t \bx_t^T$ \\
\s $t = t+1$ \\
end
\\
\hline \hline
\end{tabular}
  \caption{TD(0) with linear function approximation for LQ control}\label{LQ_TD}
\end{figure}

\subsection{Sarsa}

The cost-to-go function is used to select control actions, so the action-value function
$Q^*_t(\bx, \bu)$ is more appropriate for this purpose. The action-value function is defined as
\begin{eqnarray}\label{e:Q_original}
  Q^*_t(\bx,\bu) :=
   \inf_{\bu_{t+1}, \bu_{t+2}, \ldots} E\bigl[ c(\bx_t,\bu_t) + c(\bx_{t+1},\bu_{t+1}) +
  \ldots + c_f(\bx_f)  \big| \bx_t = \bx, \bu_t = \bu \bigr], \nonumber
\end{eqnarray}
and analogously to $V^*_t$, it can be shown that it is time
independent and can be written in the form
\begin{equation}
  Q^*(\bx,\bu) =
 \begin{pmatrix} \bx^T & \bu^T \end{pmatrix}
\begin{pmatrix}
  \Theta^*_{11} & \Theta^*_{12} \\
  \Theta^*_{21} & \Theta^*_{22}
\end{pmatrix}
 \begin{pmatrix} \bx \\ \bu \end{pmatrix}
 =
 \begin{pmatrix} \bx^T & \bu^T \end{pmatrix}
 \Theta^*
 \begin{pmatrix} \bx \\ \bu \end{pmatrix}.
\end{equation}

Note that $\Pi^*$ can be expressed by $\Theta^*$ using the relationship $V(\bx) = \min_{\bu}
Q(\bx, \bu)$:
\begin{equation}
  \Pi^* = \Theta^*_{11} - \Theta^*_{12} (\Theta^*_{22})^{-1} \Theta^*_{21}
\end{equation}
If the $t^{th}$ estimate of $Q^*$ is $Q_t(\bx,\bu) =
[\bx^T,\bu^T]^T \Theta_t [\bx^T,\bu^T]$, then the greedy control
action is given as
\begin{eqnarray}
  \bu_t &=& \arg\min_{\bu} Q_t(\bx,\bu) = -\Theta_{22}^{-1} \frac{\Theta_{21}^T + \Theta_{21}}{2} \bx_t
  = -\Theta_{22}^{-1} \Theta_{21} \bx_t,
\end{eqnarray}
where subscript $t$ of $\Theta$ has been omitted to improve readability.

The estimation error and the weight update are similar to the
state-value case:
\begin{equation}
  \delta_t =
  \begin{cases}
    c_f(\bx_t) - Q_t(\bx_{t},\bu_{t})   & \text{if $t=t_{STOP}$}, \\
    \bigl( c(\bx_{t},\bu_{t}) + Q_t(\bx_{t+1},\bu_{t+1}) \bigr) - Q_t(\bx_{t},\bu_{t}), &
    \text{otherwise},
  \end{cases}
\end{equation}
\begin{eqnarray}
  \Theta_{t+1} &=& \Theta_t + \alpha_t \cdot \delta_t \cdot \nabla_{\Theta_t} Q_t(\bx_t, \bu_t)  \label{Theta_update} \\
            &=& \Theta_t + \alpha_t \cdot \delta_t \cdot \begin{pmatrix} \bx_t \\ \bu_t \end{pmatrix}
               \begin{pmatrix} \bx_t \\ \bu_t \end{pmatrix}^T. \nonumber
\end{eqnarray}

The algorithm is summarized in Fig. \ref{LQ_sarsa}.

\begin{figure}
\begin{tabular}{l} 
\hline \hline
\\
Initialize $\bx_0$, $\bu_0$, $\Theta_0$ \\
$\b z_{0} = (\bx_{0}^T \bu_{0}^T)^T$ \\
repeat \\
\s $\bx_{t+1} = F \bx_t + G \bu_t$ \\
\s $\nu_{t+1}:=$ random noise \\
\s $\bu_{t+1} = - (\Theta_t)_{22} (\Theta_t)_{21} \bx_{t+1}+\nu_{t+1}$ \\
\s $\b z_{t+1} = (\bx_{t+1}^T \bu_{t+1}^T)^T$ \\
\s with probability $p$, \\
\s \s $\delta_t =  \bx_t^T Q^f \bx_t - \b{z}_{t}^T \Theta_t \b{z}_{t} $\\
\s \s STOP \\
\s else \\
\s \s $\delta_t =  \bu_t^T R \bu_t + \b{z}_{t+1}^T \Theta_t \b{z}_{t+1} - \b{z}_{t}^T \Theta_t \b{z}_{t} $ \\
\s $\Theta_{t+1} = \Theta_t + \alpha_t \delta_t \b{z}_t \b{z}_t^T$ \\
\s $t = t+1$ \\
end
\\
\hline \hline
\end{tabular}
  \caption{Sarsa(0) with linear function approximation for LQ control}\label{LQ_sarsa}
\end{figure}

\section{Convergence}

\begin{thm}\label{t:main}
If\/ $\Pi_0 \ge \Pi^*$, there exists an $L$ such that $\norm{F+GL}
\le 1/\sqrt{1-p}$, then there exists a series of learning rates
$\alpha_t$ such that $0< \alpha_t \leq 1/\norm{\bx_t}^4$, $\sum_t
\alpha_t = \infty$, $\sum_t \alpha_t^2 < \infty$, and it can be
computed online. For all sequences of learning rates satisfying
these requirements, Algorithm \ref{LQ_TD} converges to the optimal
policy.
\end{thm}

The proof of the theorem can be found in Appendix \ref{app:proof}.

The same line of thought can be carried over for the action-value function $Q(\bx,\bu) = (\bx^T
\bu^T)^T \Theta (\bx^T \bu^T)$, which we do not detail here, giving only the result:

\begin{thm}
If\/ $\Theta_0 \ge \Theta^*$, there exists an $L$ such that
$\norm{F+GL} \le 1/\sqrt{1-p}$, then there exists a series of
learning rates $\alpha_t$ such that $0< \alpha_t \leq
1/\norm{\bx_t}^4$, $\sum_t \alpha_t = \infty$, $\sum_t \alpha_t^2
< \infty$, and it can be computed online. For all sequences of
learning rates satisfying these requirements, Sarsa(0) with LFA
(Fig. \ref{LQ_sarsa}) converges to the optimal policy.
\end{thm}

\section{Kalman filter LQ control}

Now let us examine the case when we do not know the exact states, but we have to estimate them
from noisy observations. Consider a linear dynamical system with state $\b x_t \in \Real^n$,
control $\b u_t \in \Real^m$, observation $\b y_t \in \Real^k$, noises $\xi_t \in \Real^n$ and
$\zeta_t \in \Real^k$ (which are assumed to be uncorrelated Gaussians with covariance matrix
$\Omega^\xi$ and $\Omega^\zeta$, respectively), in discrete time $t$:
\begin{eqnarray}
  \b x_{t+1} &=& F \b x_t + G \b u_t + \b \xi_t \\
  \b y_t &=& H \b x_t + \b \zeta_t.
\end{eqnarray}
Assume that the initial state has mean $\hat{\b x}_1$, and
covariance $\Sigma_1$. Furthermore, assume that executing control
step $\b u_t$ in $\b x_t$ costs
\begin{equation}
c(\bx_t, \bu_t) := \bx_t^T Q \bx_t + \bu_t^T R \bu_t,
\end{equation}
After each time step, the process will be stopped with some fixed
probability $p$, and then the controller incurs the final cost
$c_f(\bx_f) := \bx_f^T Q^f \bx_f$.

We will show that the separation principle holds for our problem, i.e. the control law and the state
filtering can be computed independently from each other. On one hand, state estimation is independent of the
control selection method (in fact, the control could be anything, because it does not affect the estimation
error), i.e. we can estimate the state of the system by the standard Kalman filtering equations:
\begin{eqnarray}
  \hat{\bx}_{t+1} &=& F \hat{\bx}_t + G \bu_t + K_t(\by_t - H\hat{\bx}_t) \\
  K_t &=& F \Sigma_t H^T (H \Sigma_t H^T + \Omega^e)^{-1} \\
  \Sigma_{t+1} &=& \Omega^w + F \Sigma_t F^T - K_t H \Sigma_t F^T.
\end{eqnarray}
On the other hand, it is easy to show that the optimal control can be expressed as the function of
$\hat{\bx}_t$. The proof (similarly to the proof of the original separation principle) is based on the fact
that the noise and error terms appearing in the expressions are either linear and have zero mean or
quadratic and independent of $\bu$. In both cases they can be omitted. More precisely, let $W_t$ denote the
sequence $\by_1, \ldots, \by_t, \bu_1, \ldots, \bu_{t-1}$, and let $\b e_t = \bx_t - \hat{\bx}_t$. Equation
(\ref{u_t}) for the filtered case can be formulated as
\begin{eqnarray}
  \bu_t &=& \arg\min_{\bu} E \Bigl( c(\bx_t,\bu)+ V_t(F \bx_t + G \bu + \xi_t) \Big| W_t \Bigr) \\
      &=& \arg\min_{\bu} E \Bigl( \bx_t^T Q \bx_t + \bu^T R \bu + \nonumber \\
      &&(F \bx_t + G \bu + \xi_t)^T \Pi_t (F \bx_t + G \bu + \xi_t) \Big| W_t\Bigr). \nonumber
\end{eqnarray}
Using the fact that $E(\bx_t^T Q \bx_t|W_t)$ and $E(\xi_t^T \Pi_t \xi_t|W_t)$ are independent of $\bu$ and
that $E (( F \bx_t + G \bu)^T \Pi_t \xi_t | W_t) = 0$, furthermore that $\bx_t = \hat{\bx}_t + \b e_t$, we
get
\begin{eqnarray*}
  \bu_t &=& \arg\min_{\bu} E \Bigl( \bu^T R \bu + (F \hat{\bx}_t + F \b e_t + G \bu)^T \Pi_t (F \hat{\bx}_t + F \b e_t + G \bu) \Big| W_t\Bigr)
\end{eqnarray*}
Finally, we know that $E(\b e_t | W_t) = 0$, because the Kalman filter is an unbiased estimator, furthermore
$E(\b e_t^T \Pi_t \b e_t | W_t)$ is independent of $\bu$, which yields
\begin{eqnarray*}
  \bu_t &=& \arg\min_{\bu} E \Bigl( \bu^T R \bu + (F \hat{\bx}_t + G \bu)^T \Pi_t (F \hat{\bx}_t + G \bu) \Big| W_t\Bigr)\\
        &=& -(R+G^T \Pi_t G)^{-1} (G^T \Pi_t F) \hat{\bx}_t,
\end{eqnarray*}
i.e. for the computation of the greedy control action according to $V_t$ we can use the estimated state
instead of the exact one. The proof of the separation principle for SARSA(0) is quite similar and therefore
is omitted here.

The resulting algorithm using TD(0) is summarized in Fig.
\ref{kalman_TD}. The algorithm using Sarsa can be derived in a
similar manner.

\begin{figure}
\begin{tabular}{l} 
\hline \hline
\\
Initialize $\bx_0$, $\hat{\bx}_0$, $\bu_0$, $\Pi_0$, $\Sigma_0$ \\
repeat \\
\s $\bx_{t+1} = F \bx_t + G \bu_t + \xi_t$ \\
\s $\b y_t = H \b x_t + \b \zeta_t$ \\
\s $\Sigma_{t+1} = \Omega^\xi + F \Sigma_t F^T - K_t H \Sigma_t F^T$ \\
\s $K_t = F \Sigma_t H^T (H \Sigma_t H^T + \Omega^\zeta)^{-1}$ \\
\s $\hat{\bx}_{t+1} = F \hat{\bx}_t + G \bu_t + K_t(\by_t - H\hat{\bx}_t)$ \\
\s $\nu_{t+1}:=$ random noise \\
\s $\bu_{t+1} = - (R + G^T \Pi_{t+1} G)^{-1} (G^T \Pi_{t+1} F) \hat{\bx}_{t+1}+\nu_{t+1}$ \\
\s with probability $p$, \\
\s \s $\delta_t =  \hat{\bx}_t^T Q^f \hat{\bx}_t - \hat{\bx}_{t}^T \Pi_t \hat{\bx}_{t} $\\
\s \s STOP \\
\s else \\
\s \s $\delta_t =  \bu_t^T R \bu_t + \hat{\bx}_{t+1}^T \Pi_t \hat{\bx}_{t+1} - \hat{\bx}_{t}^T \Pi_t \hat{\bx}_{t} $ \\
\s $\Pi_{t+1} =  \Pi_t + \alpha_t \delta_t \hat{\bx}_t \hat{\bx}_t^T$ \\
\s $t = t+1$ \\
end
\\
\hline \hline
\end{tabular}
  \caption{Kalman filtering with TD control}\label{kalman_TD}
\end{figure}

\section{Acknowledgments}

This work was supported by the Hungarian National Science
Foundation (Grant No. T-32487). We would like to thank L{\'a}szl{\'o}
Gerencs{\'e}r for calling our attention to a mistake in the previous
version of the convergence proof.

\appendix

\section{The boundedness of $\norm{\bx_t}$}

We need several technical lemmas to show that $\norm{\bx_t}$
remains bounded for the linear-quadratic case, and also, $E
(\norm{\bx_t})$ remains bounded for the Kalman filter case. The
latter result implies that for the KF case, $\norm{\bx_t}$ remains
bounded with high probability.

For any positive semidefinite matrix $\Pi$ and any state $\bx$, we
can define the action vector which minimizes the one-step-ahead
value function:
\begin{eqnarray*}
  \bu_{greedy} &:=& \arg\min_{\bu} \Bigl( \bu^T R \bu + (F {\bx} + G \bu)^T \Pi (F {\bx} + G \bu) \Bigr)\\
        &=& -(R+G^T \Pi G)^{-1} (G^T \Pi F) {\bx}.
\end{eqnarray*}
Let
\[
    L_\Pi := -(R+G^T \Pi G)^{-1} (G^T \Pi F)
\]
denote the greedy control for matrix $\Pi$, and let
$$L^* = - (R + G^T \Pi^* G)^{-1} (G^T \Pi^* F) $$
be the optimal policy, furthermore, let $q := 1/\sqrt{1-p}$.

\begin{lem} \label{L_lemma}
If there exists an $L$ such that $\norm{F+GL} < q$, then
$\norm{F+GL^*} < q$ as well.
\end{lem}
\begin{proof}
Indirectly, suppose that $\norm{F+GL^*} \ge q$. Then for a fixed
$\bx_0$, let $\bx_t$ be the optimal trajectory
$$ \bx_{t+1} = (F + GL^*)\bx_t. $$
Then
\begin{eqnarray*}
 V^*(\bx_0) =& p\ c_f(\bx_0)  &+ (1-p) c(\bx_0, L^*\bx_0) \\
         +& (1-p)p\ c_f(\bx_1) &+ (1-p)^2 c(\bx_1, L^*\bx_1) \\
         +& (1-p)^2 p\ c_f(\bx_2) &+ (1-p)^3 c(\bx_2, L^*\bx_2) \\
         + &\ldots,
\end{eqnarray*}
\begin{eqnarray*}
   V^*(\bx_0) &\ge& p \bigl(c_f(\bx_0) + (1-p) c_f(\bx_1) + (1-p)^2 c_f(\bx_2) +\ldots \bigr) \\
     &=& p \sum (1-p)^k \bx_0^T {(F+GL^*)^k}^T Q^f (F+GL^*)^k \bx_0.
\end{eqnarray*}
We know that $Q^f$ is positive definite, so there exists an
$\epsilon$ such that $\bx^T Q^f \bx \ge \epsilon \norm{\bx}^2$,
therefore
\begin{eqnarray*}
   V^*(\bx_0) &\ge& \epsilon p \sum (1-p)^k \norm{(F+GL^*)^k \bx_0}^2.
\end{eqnarray*}
If $\bx_0$ is the eigenvector corresponding to the maximal
eigenvalue of $F+GL^*$, then $(F+GL^*) \bx_0 = \norm{F+GL^*}
\bx_0$, and so $(F+GL^*)^k \bx_0 = \norm{F+GL^*}^k \bx_0$.
Consequently,
\begin{eqnarray*}
   V^*(\bx_0) &\ge& \epsilon p \sum (1-p)^k \norm{F+GL^*}^{2k} \norm{\bx_0}^2 \\
     &\ge& \epsilon p \sum (1-p)^k \frac{1}{(1-p)^{k}} \norm{\bx_0}^2 = \infty.
\end{eqnarray*}
On the other hand, because of $\norm{F+GL} < q$, it is easy to see
that the value of following the control law $L$ from $\bx_0$ is
finite, therefore we get $V^L(\bx_0) < V^*(\bx_0)$, which is a
contradiction.
\end{proof}

\begin{lem}
For positive definite matrices $A$ and $B$, if $A \geq B$ then
$\norm{A^{-1}B} \leq 1$.
\end{lem}
\begin{proof}
Indirectly, suppose that $\norm{A^{-1}B} > 1$. Let $\lambda_{max}$
be the maximal eigenvalue of $A^{-1}B$, and $\b v$ be a
corresponding eigenvector.
\[
  A^{-1}B \b v = \lambda_{max} \b v,
\]
and according to the indirect assumption,
\[
  \lambda_{max} = \norm{A^{-1}B} > 1.
\]
$A\geq B$ means that for each $\bx$, $\bx^T A \bx \geq \bx^T B
\bx$, so this holds specifically for $\bx = A^{-1}B \b v =
\lambda_{max} \b v$, too. So, on one hand,
\[
   (\lambda_{max} \b v)^T B (\lambda_{max} \b v) = \lambda_{max}^2
   {\b v}^T B \b v > {\b v}^T B \b v,
\]
and on the other hand,
\[
   (\lambda_{max} \b v)^T A (\lambda_{max} \b v) = (A^{-1}B  \b v)^T A (A^{-1}B \b
   v) = {\b v}^T (B A^{-1} B) \b v,
\]
so,
\[
    {\b v}^T (B A^{-1} B) \b v > {\b v}^T B \b v,
\]

However, from $A \geq B$, $A^{-1} \leq B^{-1}$. Multiplying this
with $B$ from both sides, we get $BA^{-1}B \leq B$, which is a
contradiction.
\end{proof}

\begin{lem} \label{L_stab_lemma}
If there exists an $L$ such that $\norm{F+GL} < q$ then for any
$\Pi$ such that $\Pi \geq \Pi^*$,  $\norm{F+GL_\Pi} < q$, too.
\end{lem}
\begin{proof}
We will apply the Woodbury identity \cite{petersen05matrix},
stating that for positive definite matrices $R$ and $\Pi$,
\[
  (R + G^T \Pi G)^{-1} G^T \Pi = R^{-1} G^T (GR^{-1}G^T +
  \Pi^{-1})^{-1}
\]
Consequently,
\begin{eqnarray*}
F+GL_\Pi &=& F - G (R+G^T \Pi G)^{-1} (G^T \Pi F) \\
    &=& F - \Bigl( G R^{-1} G^T \Bigr) \Bigl( (GR^{-1}G^T +
  \Pi^{-1})^{-1}\Bigr) F.
\end{eqnarray*}
Let
\begin{eqnarray*}
  U_\Pi &:=& I - \Bigl( G R^{-1} G^T \Bigr) \Bigl( GR^{-1}G^T +
  \Pi^{-1}\Bigr)^{-1} \\
      &=& \Pi^{-1} \Bigl( GR^{-1}G^T +
  \Pi^{-1}\Bigr)^{-1}
\end{eqnarray*}
and
\begin{eqnarray*}
  U^* &:=& I - \Bigl( G R^{-1} G^T \Bigr) \Bigl( GR^{-1}G^T +
  (\Pi^*)^{-1}\Bigr)^{-1} \\
      &=& (\Pi^*)^{-1} \Bigl( GR^{-1}G^T +
  (\Pi^*)^{-1}\Bigr)^{-1}
\end{eqnarray*}
Both matrices are positive definite, because they are the product
of positive definite matrices. With these notations, $F+GL_\Pi =
U_\Pi F$ and $F+GL^* = U^* F$.

It is easy to show that $U_\Pi \leq U^*$ exploiting the fact that
$\Pi \geq \Pi^*$ and several well-known properties of matrix
inequalities: if $A \geq B$ and $C$ is positive semidefinite, then
$-A \leq -B$, $A^{-1} \leq B^{-1}$, $A+C \geq B+C$, $A\cdot C \geq
B \cdot C$.

From Lemma~\ref{L_lemma} we know that $\norm{U^*F} =
\norm{F+GL^*}< q$, and from the previous lemma we know that
$\norm{U_\pi (U^*)^{-1}} \leq 1$, so
\[
  \norm{F+GL_\Pi} = \norm{U_\Pi F} = \norm{U_\Pi (U^*)^{-1} U^* F}
  \leq \norm{U_\Pi (U^*)^{-1}} \norm{ U^* F} \leq 1 \cdot q
\]
\end{proof}

\begin{cor}
If there exists an $L$ such that $\norm{F+GL} \le q$, then the
state sequence generated by the noise-free LQ equations is
bounded, i.e., there exists $M\in \Real$ such that $\norm{\bx_t}
\leq M$.
\end{cor}
\begin{proof}
This is a simple corollary of the previous lemma: in each step we
use a greedy control law $L_t$, so
\[
  \norm{\bx_{t+1}} = \norm{(F+GL_t) \bx_t} \leq q \norm{\bx_t}
\]
\end{proof}

\begin{cor}
If there exists an $L$ such that $\norm{F+GL} \le q$, then the
state sequence generated by the Kalman-filter equations is bounded
with high probability, i.e., for any $e>0$, there exists $M\in
\Real$ such that $\norm{\bx_t} \leq M$ with probability
$1-\epsilon$.
\end{cor}
\begin{proof}
\begin{eqnarray*}
  E \norm{\bx_{t+1}} &=& E\norm{(F+GL_t) \bx_t + \xi_t} \leq \sqrt{E\norm{(F+GL_t)
  \bx_t}+ \Omega_\xi} \\
    &\leq& \sqrt{q E\norm{\bx_t} + \Omega_\xi},
\end{eqnarray*}
so there exists a bound $M'$ such that $E \norm{\bx_t} \leq M'$.
From Markov's inequality,
\[
  \Pr ( \norm{\bx_t} > M'/e) < e,
\]
therefore, $M = M'/e$ satisfies our requirements.
\end{proof}

\section{The proof of the main theorem} \label{app:proof}

We will use the following lemma:

\begin{lem} \label{lemma}
Let $J$ be a differentiable function, bounded from below by $J^*$,
and let $\nabla J$ be Lipschitz-continuous. Suppose the weight
sequence $w_t$ satisfies
  $$ w_{t+1} = w_t + \alpha_t b_t $$
for random vectors $b_t$ independent of $w_{t+1}, w_{t+2},
\ldots$, and $b_t$ is a descent direction for $J$, i.e.
$E(b_t|w_t)^T \nabla J(w_t) \le - \delta(\epsilon) < 0$ whenever
$J(w_t)
> J^* + \epsilon$. Suppose also that
 $$ E(\|{b_t}\|^2 | w_t) \le K_1 J(w_t) + K_2 E(b_t|w_t) ^T \nabla J(w_t) + K_3 $$
and finally that the constants $\alpha_t$ satisfy $\alpha_t > 0$,
$\sum_t \alpha_t = \infty$, $\sum_t \alpha_t^2 < \infty$. Then
$J(w_t) \to J^*$ with probability 1.
\end{lem}

In our case, the weight vectors are $n\times n$ dimensional, with
$w_{n\cdot i +j} := \Pi_{ij} $. For the sake of simplicity, we
denote this by $w_{(ij)}$. Let $w^*$ be the weight vector
corresponding to the optimal value function, and let
$$ J(w) = \frac{1}{2}\norm{w - w^*}^2. $$

\begin{thm}[Theorem \ref{t:main}]
If\/ $\Pi_0 \ge \Pi^*$, there exists an $L$ such that $\norm{F+GL}
\le q$, then there exists a series of learning rates $\alpha_t$
such that $0< \alpha_t \leq 1/\norm{\bx_t}^4$, $\sum_t \alpha_t =
\infty$, $\sum_t \alpha_t^2 < \infty$, and it can be computed
online. For all sequences of learning rates satisfying these
requirements, Algorithm \ref{LQ_TD} converges to the optimal
policy.
\end{thm}

\begin{proof}

First of all, we prove the existence of a suitable learning rate
sequence. Let $\alpha'_t$ be a sequence of learning rates that
satisfy two of the requirements, $\sum_t \alpha_t = \infty$ and
$\sum_t \alpha_t^2 < \infty$. Fix a probability $0<e<1$. By the
previous lemma, there exists a bound $M$ such that $\norm{\bx_t}
\leq M$ with probability $1-e$. The learning rates
\[
  \alpha_t := \min\{\alpha'_t, 1/\norm{\bx_t}^4\}
\]
will be satisfactory, and can be computed on the fly. The first
and third requirements are trivially satisfied, so we only have to
show that $\sum_t \alpha_t = \infty$. Consider the index set $H =
\{ t : \alpha'_t \leq 1/M^4 \} \cup \{t:  \alpha'_t \leq
1/\norm{\bx_t}^4 \}$. By the first condition only finitely many
indices are excluded. The second condition excludes indices with
$1/M^4 < \alpha'_t < 1/\norm{\bx_t}^4$, which happens at most with
probability $e$. However,
\[
  \sum_t \alpha_t \geq \sum_{t\in H} \alpha_t = \sum_{t\in H}
  \alpha'_t = \infty.
\]
The last equality holds, because if we take a divergent sum of
nonnegative terms, and exclude finitely many terms or an index set
with density less than 1, then the remaining subseries will remain
divergent.

An update step of the algorithm is $\alpha_t \delta_t \bx_t
\bx_t^T$. To make the proof simpler, we decompose this into a step
size $\alpha'_t$ and a direction vector $(\alpha_t/\alpha'_t)
\delta_t \bx_t \bx_t^T$. Denote the scaling factor by
\[
A_t := \alpha_t/\alpha'_t = \min \{ 1, 1/(\alpha'_t
\norm{\bx_t}^4)\}
\]
Clearly, $A_t \leq 1$. In fact, it will be one most of the time,
and will damp only the samples that are too big.

We will show that
$$
  b_t =  A_t \delta_t \bx_t \bx_t^T
$$
is a descent direction for every $t$.

\begin{eqnarray*}
  E(b_t | w_t)^T \nabla J(w_t) &=& A_t E(\delta_t | w_t) \bx_t \bx_t^T (w_t - w^*) \\
  &=& A_t E(\delta_t | w_t) \bx_t^T (\Pi_t - \Pi^*) \bx_t \\
  &=&  A_t E(\delta_t | w_t) (V_t(\bx_t) - V^*(\bx_t)).
\end{eqnarray*}

For the sake of simplicity, from now on we do not note the
dependence on $w_t$ explicitly.

We will show that for all $t$, $E(\Pi_{t}) > 0$, $E(\Pi_{t-1}) >
E(\Pi_{t})$ and $E(\delta_t) \le -p \bx_t^T (\Pi_t-\Pi^*) \bx_t$.
We proceed by induction.

\underline{$\bullet$ $t=0$.} $\Pi_0 > \Pi^*$ holds by assumption.

\vspace{3mm} \underline{$\bullet$ Induction step part 1:
$E(\delta_t) \le -p\ \bx_t^T (\Pi_t-\Pi^*) \bx_t$.}

\noindent Recall that
\begin{eqnarray}
  \bu_t &=& \arg\min_{\bu} \Bigl( c(\bx_t,\bu)+ V_t(F \bx_t + G \bu) \Bigr) \\
      &=& L_t \bx_t,  \nonumber
\end{eqnarray}
where
$$L_t = - (R + G^T \Pi_{t} G)^{-1} (G^T \Pi_{t} F) $$
is the greedy control law with respect to $V_t$. Clearly, by the
definition of $L_t$,
$$
  c(\bx_t,L_t \bx_t)+ V_t(F \bx_t + G L_t \bx_t) \le c(\bx_t, L^* \bx_t)+ V_t(F \bx_t + G L^*
  \bx_t).
$$
This yields
\begin{eqnarray}
  E(\delta_t) &=& p\ c_f(\bx_t) + (1-p) \bigl( c(\bx_t, L_t \bx_t) + V_t(F \bx_t + G L_t \bx_t)
   \bigr) - V_t(\bx_t) \label{e:th_8} \\
  &\le& p\ c_f(\bx_t) + (1-p) \bigl( c(\bx_t, L^* \bx_t) + V_t(F \bx_t + G L^* \bx_t)\bigr) -
  V_t(\bx_t). \nonumber
\end{eqnarray}
We know that the optimal value function satisfies the fixed-point
equation
\begin{equation}
 0 =  \Bigl( p\ c_f(\bx_t) + (1-p) \bigl( c(\bx_t, L^* \bx_t) + V^*(F \bx_t + G L^*
 \bx_t)\bigr)  \Bigr) - V^*(\bx_t). \label{e:th_9}
\end{equation}
Subtracting this from Eq.~(\ref{e:th_8}), we get
\begin{eqnarray}
  E(\delta_t)  &\le&  (1-p) \bigl( V_t(F \bx_t + G L^* \bx_t) - V^*(F \bx_t + G L^* \bx_t)
  \bigr) \label{e:th_10}\\
  && - (V_t(\bx_t) - V^*(\bx_t)). \nonumber \\
  &=& (1-p) \bx_t^T (F+GL^*)^T (\Pi_t -\Pi^*) (F+GL^*) \bx_t \\
    &&- \bx_t^T (\Pi_t -\Pi^*) \bx_t.
\end{eqnarray}
Let $\epsilon_1 = \epsilon_1(p) := 1/(1-p) - \norm{F+GL^*}^2 >0$.
Inequality (\ref{e:th_10}) implies
\begin{eqnarray}
  E(\delta_t)  &\le&  (1-p) ( \frac{1}{1-p} - \epsilon_1(p)) \bx_t^T (\Pi_t -\Pi^*) \bx_t - \bx_t^T (\Pi_t -\Pi^*) \bx_t \\
   &=& - (1-p)\epsilon_1(p) \bx_t^T (\Pi_t -\Pi^*) \bx_t. \\
   &=& - \epsilon_2(p) \bx_t^T (\Pi_t -\Pi^*) \bx_t,
\end{eqnarray}
where we defined $\epsilon_2(p) = (1-p)\epsilon_1(p)$

\vspace{3mm} \underline{$\bullet$ Induction step part 2:
$E(\Pi_{t+1}) > \Pi^* $.}
\begin{eqnarray}
  E(\delta_t) &=& p\ c_f(\bx_t) + (1-p) \bigl( c(\bx_t, L_t \bx_t) + V_t(F \bx_t + G L_t \bx_t)
    \bigr) - V_t(\bx_t) \\
  &\ge& p\ c_f(\bx_t) + (1-p) \bigl( c(\bx_t, L_t \bx_t) + V^*(F \bx_t + G L_t \bx_t)\bigr) -
    V_t(\bx_t). \nonumber
\end{eqnarray}
Subtracting eq. \ref{e:th_9}, we get
\begin{eqnarray}
  E(\delta_t) &\ge& (1-p) \Bigl(\bigl( c(\bx_t, L_t \bx_t) + V^*(F \bx_t + G L_t \bx_t)\bigr) \\
    &&- \bigl( c(\bx_t, L^* \bx_t) + V^*(F \bx_t + G L^* \bx_t)\bigr)\Bigr) + V^*(\bx_t) -
    V_t(\bx_t) \nonumber \\
    &\ge& V^*(\bx_t) - V_t(\bx_t) \ge - \norm{\Pi_t - \Pi^*} \norm{\bx_t}^2. \nonumber
\end{eqnarray}
Therefore
\begin{eqnarray}
  E(\Pi_{t+1}) - \Pi^* &\ge& \Pi_t + \alpha'_t A_t E(\delta_t) \bx_t \bx_t^T - \Pi^* \\
   &\ge& (\Pi_t - \Pi^*) - \alpha_t \norm{\bx_t}^4 \norm{\Pi_t - \Pi^*}  I \\
   &\ge& (\Pi_t - \Pi^*) - \norm{\Pi_t - \Pi^*}  I > 0.
\end{eqnarray}
\vspace{3mm}

\underline{$\bullet$ Induction step part 3: $\Pi_t > E(\Pi_{t+1})
$.}
\begin{eqnarray}
  \Pi_{t} - E(\Pi_{t+1}) &=& -\alpha'_t A_t E(\delta_t) \bx_t \bx_t^T \ge \alpha_t \epsilon_2(p) \bx_t^T (\Pi_t -\Pi^*) \bx_t \cdot \bx_t
  \bx_t^T,
\end{eqnarray}
but $\alpha'_t \epsilon_2(p)>0$, $\bx_t^T (\Pi_t -\Pi^*) \bx_t >0$
and $\bx_t \bx_t^T >0$, so their product is positive as
well.\vspace{3mm}

\noindent The induction is therefore complete. \vspace{3mm}

\noindent We finish the proof by showing that the assumptions of
Lemma \ref{lemma} hold:\vspace{3mm}

\underline{ $b_t$ is a descent direction.} Clearly, if $J(w_t) \ge
\epsilon$, then $\norm{\Pi_t - \Pi^*} \ge \epsilon_3(\epsilon)$,
but $\Pi_t - \Pi^*$ is positive definite, so $\Pi_t - \Pi^* \ge
\epsilon_3(\epsilon) I$.
\begin{eqnarray*}
  E(b_t | w_t)^T \nabla J(w_t) &=&  A_t E(\delta_t | w_t) (V_t(x_t) - V^*(x_t)) \\
    &\le& - \epsilon_2(p) A_t \bx_t^T (\Pi_t -\Pi^*) \bx_t \cdot \bx_t^T (\Pi_t -\Pi^*) \bx_t\\
    &\le& - \epsilon_2 \epsilon_3^2 A_t \norm{\bx_t}^4 \\
    &\le& - \epsilon_2 \epsilon_3^2 \min\{ \norm{\bx_t}^4 , 1/\alpha'_t \}
\end{eqnarray*}\vspace{3mm}

\underline{ $E(\norm{b_t}^2|w_t)$ is bounded.} $|E(\delta_t)| \le
|\bx_t^T (\Pi_t-\Pi^*) \bx_t|$. Therefore
\begin{eqnarray*}
  E(\norm{b_t}^2|w_t) &\le& |A_t|^2 |E(\delta_t)|^2 \norm{\bx_t}^2 \\
    &\le& \norm{\Pi_t - \Pi^*}^2 \cdot \min\{ 1, 1/(\alpha'^2_t \norm{\bx_t}^8 ) \} \cdot \norm{\bx_t}^6\\
    &\le& \norm{\Pi_t - \Pi^*}^2 \cdot \min\{ \norm{\bx_t}^6 , 1 /(\alpha'^2_t \norm{\bx_t}^2 ) \} \\
  &\le& K \cdot J(w_t) .
\end{eqnarray*}

Consequently, The assumptions of lemma \ref{lemma} hold, so the
algorithm converges to the optimal value function with probability
1.
\end{proof}

%
%

\bibliographystyle{amsplain}

\end{document}